%% file: main.tex
\documentclass{new_tlp}

\usepackage{mathptmx}

\usepackage{pgfplots}
\usepackage{tikz}
\usepackage{tikz-qtree}
\usepackage{amsmath}
\usepackage{amssymb}
\usepackage{hyperref}
\usetikzlibrary{arrows,automata,positioning,fit,calc}

\usepackage{graphics}
\usepackage[ruled,vlined]{algorithm2e}

\input{macros.tex}

  \title[Theory and Practice of Logic Programming]{An Application of a Runtime Epistemic Probabilistic Event Calculus to Decision-making in e-Health Systems}

  \author[D'Asaro et al.]
         {FABIO AURELIO D'ASARO\\
         Ethos Group,\\ 
         University of Verona, Italy\\
         \email{fabioaurelio.dasaro@univr.it}
         \and
         LUCA RAGGIOLI\\
         Department of Electrical Engineering and Information Technologies,\\ 
         University of Naples Federico II, Italy\\
         \email{luca.raggioli@manchester.ac.uk}
         \and
         SALIM MALEK\\
         Department of Electrical Engineering and Information Technologies,\\ 
         University of Naples Federico II, Italy\\
         \email{salim.malek@unibz.it}
         \and
         MARCO GRAZIOSO, SILVIA ROSSI\\
         Department of Electrical Engineering and Information Technologies,\\ 
         University of Naples Federico II, Italy\\
         \email{\{marco.grazioso,silrossi\}@unina.it}\\
         }
\jdate{March 2003}
\pubyear{2003}
\pagerange{\pageref{firstpage}--\pageref{lastpage}}
\doi{S1471068401001193}

\newtheorem{scenario}{Scenario}[section]
\newtheorem{proposition}{Proposition}[section]
\pgfplotsset{compat=1.17}

\begin{document}
\sloppy

\label{firstpage}

\maketitle

  \begin{abstract}
    We present and discuss a runtime architecture that integrates sensorial data and classifiers with a logic-based decision-making system in the context of an e-Health system for the rehabilitation of children with neuromotor disorders. In this application, children perform a rehabilitation task in the form of games. The main aim of the system is to derive a set of parameters the child's current level of cognitive and behavioral performance (e.g., engagement, attention, task accuracy) from the available sensors and classifiers (e.g., eye trackers, motion sensors, emotion recognition techniques) and take decisions accordingly. These decisions are typically aimed at improving the child's performance by triggering appropriate re-engagement stimuli when their attention is low, by changing the game or making it more difficult when the child is losing interest in the task as it is too easy. Alongside state-of-the-art techniques for emotion recognition and head pose estimation, we use a runtime variant of a probabilistic and epistemic logic programming dialect of the Event Calculus, known as the Epistemic Probabilistic Event Calculus. In particular, the probabilistic component of this symbolic framework allows for a natural interface with the machine learning techniques. We overview the architecture and its components, and show some of its characteristics through a discussion of a running example and experiments. Under consideration for publication in Theory and Practice of Logic Programming (TPLP). 
  \end{abstract}


\section{Introduction}

In this paper, we present and discuss an architecture for integrating and reasoning about sensors in the context of the \textit{AVATEA} project (\textit{Advanced Virtual Adaptive Technologies e-hEAlth}). The final goal of the project is to design and implement an integrated system to support the rehabilitation process of children with \textit{Development Coordination Disorders} (\textit{DCD}s). The system needs to integrate and control several components, including an adjustable seat, various types of sensors, and an interactive visual interface to perform rehabilitation exercises in the form of games (such games are sometimes called \textit{exergames}, presented by Vernadakis et al. \shortcite{Vernadakis2015}). One of the main goals of the project is to automatize the therapeutic task, ideally without making it less effective. To this end, we employ a logic-based AI engine that collects data from the environment and decides what strategy may be implemented to make the video game as challenging as possible while keeping the child engaged and attentive. For example, the engine may decide to emit a sound if the visual attention of the user is detected to be low, or increase the video game's difficulty level if the sensors seem to indicate that the child is bored as s/he is performing the task effortlessly.

\begin{figure}
    \includegraphics[width=\textwidth]{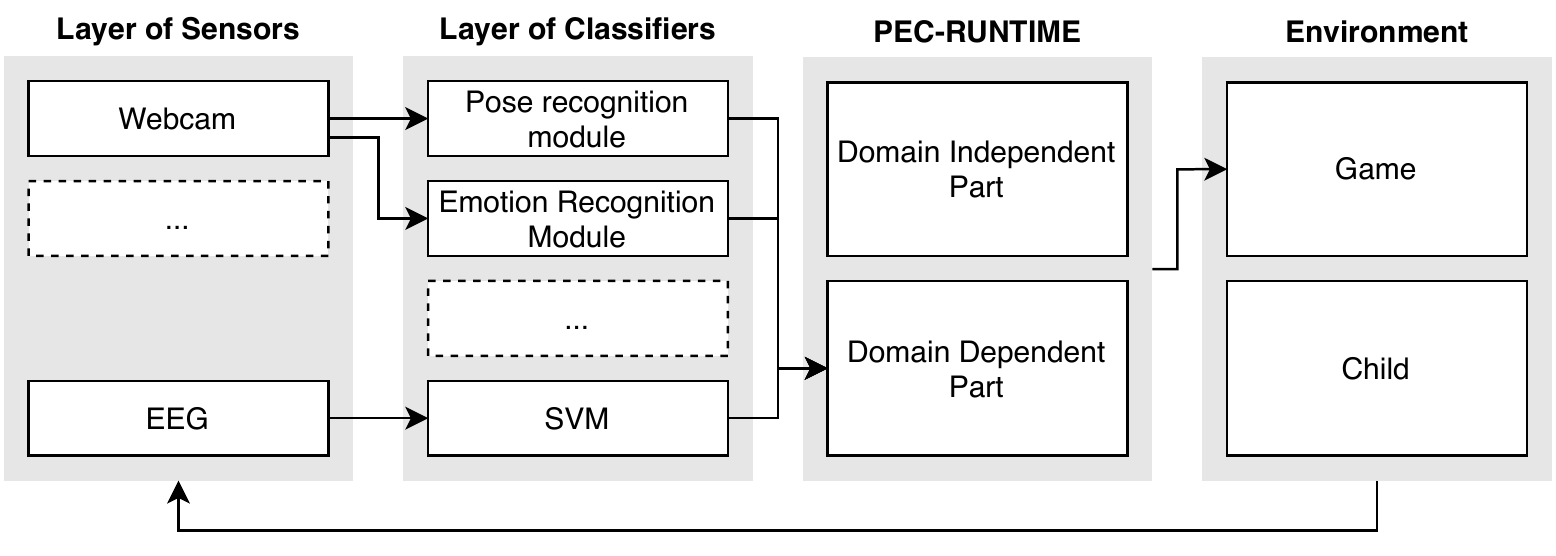}
    \caption{The architecture of the AVATEA system. Several classifiers are applied to a stream of data from different sensors. Note that a single sensor may produce data that is then fed into two or more classifiers (e.g. in the case of the webcam). Timestamped output from the classifiers is fed into PEC-RUNTIME, the logical core of the architecture. PEC-RUNTIME processes this information together with some domain independent axioms and outputs its decision to the environment (the game, in this case).}
    \label{fig:architecture}
\end{figure}

Although this architecture has been specifically designed for the rehabilitation of children with neuro-motor disorders, it can be more generally applied to any task requiring a system to take runtime decisions according to a stream of sensorial data. It mainly consists of four modules communicating with each other (see Figure \ref{fig:architecture}). Layers of sensors and classifiers (e.g., a webcam paired with head-pose and emotion recognition algorithms) collect and process information about the current state of the environment and the child undergoing the rehabilitation process. This information is fed to a probabilistic logic programming system, a modified version of the \textit{Epistemic Probabilistic Event Calculus} (EPEC for short, first introduced by D'Asaro et al. \shortcite{dasaro2017lpnmr} as a non-epistemic framework under the name PEC and extended in \cite{dasaro2019phd} to the epistemic case). To guarantee good performance at runtime, in this work, we present a novel implementation of EPEC, dubbed \textit{PEC-RUNTIME}, which implements a form of \textit{progression} \cite{lin1997howtoprogress}. PEC-RUNTIME processes the sensor data and takes decisions according to a predefined strategy. Finally, the gaming platform actuates these decisions and communicates the new state of the game to the layer of sensors.

EPEC (and consequently PEC-RUNTIME) is particularly well-suited for this task as its probabilistic nature facilitates the communication with probabilistic machine learning-based classifiers. As it is a symbolic framework, it can also be used to provide accurate human-understandable reports of the user activity at the end of each therapeutic session. Typical feedback includes information about attention levels of the child (e.g., ``The user was highly engaged for 36 seconds during the session''), justification for the decisions taken (e.g., ``Switched to higher difficulty level as the user has been carrying out the exercise correctly for the last 10 seconds''), as well as a complete report of what happened throughout the therapeutic session (e.g., ``The user was not visible at 15:43:23'', ``Visual stimulus presented at 15:43:24'', etc.) and a series of graphs for accurate tracking of user activity.

This paper extends \cite{dasaro2019woa} and is organized as follows. In Section \ref{sec:OverviewOfTheLanguage}, we overview the logical part of our architecture, EPEC, and show some of its features through the discussion of a toy example. In Section \ref{sec:ARuntimeEPEC}, we discuss how we adapted the original ASP-based inference mechanism of EPEC in order to work at runtime. In Section \ref{sec:Architecture}, we briefly describe the layers of sensors and classifiers. 
In Section \ref{sec:Examples}, we demonstrate our approach in our use-case and provide a systematic empirical evaluation. In Section \ref{sec:RelatedWork} we discuss related work, in particular probabilistic logic programming languages for reasoning about actions and gamification techniques. Some final remarks about directions for further work conclude the paper in Section \ref{sec:Conlcusion}.

\section{Overview of the language}\label{sec:OverviewOfTheLanguage}
    In this section, we overview a subset of EPEC\footnote{The fully fledged framework was introduced in \cite{dasaro2019phd}, and has some minor differences from the version presented here.} that is relevant to our application, and show how it can be used to represent a domain. Similarly to other logical languages for reasoning about actions, EPEC models a domain as a collection of \textit{fluents} and \textit{actions}. Moreover, EPEC represents time explicitly through \textit{instants}. Fluents, actions and instants constitute the principal sorts of EPEC. Actions are further sub-divided into \textit{agent actions} (under the control of the agent being modeled) and \textit{environmental actions} (performed by the environment). Fluents can take value in a set of \textit{values}. Domain-specific theories can be designed using EPEC propositions. As an illustration, we use the following toy domain (inspired by the AVATEA use-case) as our running example (but you can see e.g. \cite{acciaro2021predicting} for other use-cases):
    
    \begin{scenario}\label{ex:RunningExample} A child undergoes a simple attention test in which her level of visual attention is measured. The task consists in tracking a moving object on a screen. Therapists agree that when the child is not looking at the object, there is $90\%$ chance this is due to lack of attention, and that when the child is looking at the object there is $100\%$ chance s/he is paying attention to the task. A webcam is used to track her gaze. Readings from the webcam are only partly reliable due to hardware limitations, and are produced every second. These readings are then piped into a specialized classifier which decides whether the child is looking at the object on the screen or not. The classifier also outputs a confidence level for its classification. When the child is not looking at the screen, it may be useful to play a sound in order to re-engage the child. Two seconds after the start of the experiment, the child has been detected to be tracking the object on the screen with confidence levels $0.25$ and $0.13$. What is the level of attention of the child at instant $2$? Should the sound be played at time $2$?''. \end{scenario}
    
    We let the set of instants in our timeline be $\{ 0, 1, 2 \}$, whose elements can be naturally interpreted as seconds since the start of the experiment.
    
    \subsection{Syntax overview}
    The only fluent in our example scenario is $\Attention$. It is \textit{boolean}, i.e., it may take values $\True$ and $\False$. In EPEC, this is written
    \begin{equation}\label{eq:exampleVProp}
    \Attention \takesValues \{ \True, \False \}
    \end{equation}
    
    In the example, we consider the agent to be an automated system that is responsible for triggering a re-engagement strategy when necessary. On the other hand, the child is considered part of the environment. Therefore, the actions are $\PlaySound$ and $\TrackObject$, where $\PlaySound$ is the agent action of playing a sound and $\TrackObject$ is the environmental action which corresponds to the child tracking the object with her eyes. The effects of \textit{not} tracking the object are defined by the following proposition:
    \begin{equation}\label{eq:exampleCProp1}
        \neg \TrackObject \causesOneOf \{ (\{\neg \Attention \}, 0.9), (\emptyset,0.1) \}
    \end{equation}
    
    Looking at the object is never considered to be due to chance, therefore we also include the following proposition:
    \begin{equation}\label{eq:exampleCProp2}
        \TrackObject \causesOneOf \{ (\{\Attention \}, 1) \}
    \end{equation}
    
    For what regards action $\PlaySound$ under the control of the agent, we want to formalize the \textit{conditional plan} stating that this action must be performed at instant $2$ if the child is believed to be distracted, i.e., if the probability of $\Attention$ has fallen below a predefined threshold, say $0.1$. This can be written in EPEC as:
    \begin{equation}\label{eq:examplePProp}
        \PlaySound \performedAt 2 \ifBelieves (\Attention, [0, 0.1])
    \end{equation}
    
    We consider the child to be initially paying full attention to the task. This translates to:
    \begin{equation}\label{eq:exampleIProp}
        \initiallyOneOf \{ (\{\Attention\},1) \}
    \end{equation}
    
    Finally, the action $\TrackObject$ occurs twice at instants $0$ and $1$ with confidence $0.25$ and $0.13$ respectively. This translates to the following propositions:
    \begin{gather}
        \TrackObject \occursAt 0 \withProb 0.25\label{eq:exampleOProp1}\\
        \TrackObject \occursAt 1 \withProb 0.13\label{eq:exampleOProp2}
    \end{gather}
    In EPEC, Proposition (\ref{eq:exampleVProp}) is known as a \textit{v-proposition} (\textit{v} for ``value''), Propositions (\ref{eq:exampleCProp1}) and (\ref{eq:exampleCProp2}) are known as \textit{c-propositions} (\textit{c} for ``causes''), Proposition (\ref{eq:examplePProp}) is known as a \textit{p-proposition} (\textit{p} for ``performs''), Proposition (\ref{eq:exampleIProp}) is known as an \textit{i-proposition} (\textit{i} for ``initially''), Propositions (\ref{eq:exampleOProp1}) and (\ref{eq:exampleOProp2}) are known as \textit{o-propositions} (\textit{o} for occurs). Propositions (\ref{eq:examplePProp}), (\ref{eq:exampleOProp1}) and (\ref{eq:exampleOProp2}) are known as the \textit{narrative} part of the domain. 
    
    The set of propositions $\{\eqref{eq:exampleVProp},\dots, \eqref{eq:exampleOProp2}\}$ constitutes what in EPEC is called a \textit{domain description} and is usually denoted by $\mathcal{D}$ (with appropriate subscripts and/or supercripts). The narrative part of a domain is denoted by $\narr(\mathcal{D})$. A domain description is given meaning using a bespoke semantics.
    
\subsection{Semantics overview}\label{sec:semantics}
    
    The semantics of EPEC enumerates all the possible evolutions (\textit{worlds} in the terminology of EPEC) of the environment being modeled, starting from the initial state. It then applies a series of standard reasoning about actions principles (e.g. persistence of fluents, closed world assumption for actions) to filter out those evolutions of the environment that do not make intuitive sense with respect to the given domain. The remaining worlds, dubbed \textit{well-behaved worlds} to indicate that they are meaningful with respect to the given domain description, are assigned a probability value. Some well-behaved worlds for our example domain can be depicted as follows:
    \begin{center}
    		\begin{tikzpicture}[scale=0.9]
    		\draw[<-] (12,0) -- (-2,0) node[anchor=east] {$ W_1 $};
    		\draw (0,-0.1) -- (0,0.1) node[anchor=south,text width=2.5cm,align=center] {\small$\{ \Attention,$ $\TrackObject,$ $\neg \PlaySound \}$};
    		\draw (5,-0.1) -- (5,0.1) node[anchor=south,text width=2.5cm,align=center] {\small$\{ \Attention,$ $\neg \TrackObject,$ $\neg \PlaySound \}$};
    		\draw (10,-0.1) -- (10,0.1) node[anchor=south,text width=2.5cm,align=center] {\small$\{ \neg \Attention,$ $\neg \TrackObject,$ $\neg \PlaySound \}$};
    		\node at (0,-0.3) {{\small {$0$}}};
    		\node at (5,-0.3) {{\small {$1$}}};
    		\node at (10,-0.3) {{\small {$2$}}};
    		\node at (12,-0.3) {{\small {$I$}}};
    		\end{tikzpicture}
    		\\[1em]
    		\begin{tikzpicture}[scale=0.9]
    		\draw[<-] (12,0) -- (-2,0) node[anchor=east] {$ W_2 $};
    		\draw (0,-0.1) -- (0,0.1) node[anchor=south,text width=2.5cm,align=center] {\small$\{ \Attention,$ $\neg \TrackObject,$ $\neg \PlaySound \}$};
    		\draw (5,-0.1) -- (5,0.1) node[anchor=south,text width=2.5cm,align=center] {\small$\{ \Attention,$ $\neg \TrackObject,$ $\neg \PlaySound \}$};
    		\draw (10,-0.1) -- (10,0.1) node[anchor=south,text width=2.5cm,align=center] {\small$\{  \Attention,$ $\neg \TrackObject,$ $\neg \PlaySound \}$};
    		\node at (0,-0.3) {{\small {$0$}}};
    		\node at (5,-0.3) {{\small {$1$}}};
    		\node at (10,-0.3) {{\small {$2$}}};
    		\node at (12,-0.3) {{\small {$I$}}};
    		\end{tikzpicture}
    \end{center}
    
    World $W_1$ represents a possible evolution of the environment starting from a state in which the child is initially (at instant $0$) fully attentive and tracking the object with her eyes. At instant $1$, the child is still paying attention to the task, but he/she is no longer tracking the object. At instant $2$, the child is distracted and again not tracking the object. The sound is never played. EPEC's semantics assigns a probability of $0.19575$ to this world, as this results from the product of $0.25$ (due to proposition (\ref{eq:exampleOProp1}) and $\TrackObject$ occurring at $0$ in $W_1$), $1-0.13$ (due to proposition (\ref{eq:exampleOProp2}) and $\TrackObject$ not occurring at $1$ in $W_1$) and $0.9$ (due to proposition $\ref{eq:exampleCProp1}$ and since not tracking the object at instant $1$ caused $\Attention$ not to hold at instant $2$). The intuitive meaning of world $W_2$ and its probability can be figured similarly. In $W_2$ the child is always attentive but never tracking the object, the sound is never played, and its probability is $ 0.006525 $.
    
    There are $8$ well-behaved worlds with respect to our example domain, and they are such that their probabilities sum up to $1$. Formal results about the semantics of EPEC guarantee that this is always the case for any domain description and the corresponding set of well-behaved worlds.
    
    We now want to address the question ``\textit{What is the level of attention of the child at instant $2$?}''. EPEC first needs to \textit{query} the domain description about the probability of fluent $\Attention$ at instant $2$. In EPEC, queries are expressed in the form of an \textit{i-formula} (i.e., a \textit{formula} with instants attached). In our case, the query has the form $Q_1=\ifor{\Attention}{2}$, but more complicated queries are also possible, e.g. $Q_2=(\ifor{\neg Attention}{1} \vee \ifor{\neg \TrackObject}{1}) \wedge \ifor{\Attention}{2}$. We say that i-formula $Q_1$ \textit{has instant $2$} (as this is the only instant appearing in the formula) and that the i-formula $Q_2$ has instants $>0$ (as all the instants appearing in the formula are strictly greater than $0$). Answering query $\ifor{\Attention}{2}$ amounts to summing the probabilities of those well-behaved worlds in which the i-formula is true, i.e., such that $\Attention$ is true at instant $2$. This results in $0.158275$, and can also be written
    \[
        \mathcal{D} \mmodels \ifor{\Attention}{2} \holdsWithProb 0.158275
    \]
    
    Note that this value falls outside the interval $[0, 0.1]$ in the precondition of Proposition (\ref{eq:examplePProp}). Therefore, EPEC deduces that the answer to our last question in Scenario \ref{ex:RunningExample} (``\textit{Should the sound be played at time $2$?}'') is \textit{no}. Then, Proposition (\ref{eq:examplePProp}) does not \textit{fire} and the action $\PlaySound$, under the control of the agent, is not played in any of the well-behaved worlds.
    
    \subsection{Simulating epistemic actions via c-propositions and o-propositions}
    The reader may have noticed that our target application domain mainly deals with sensors. In its original formulation \cite{dasaro2019phd}, epistemic modeling is reserved to an additional, type of propositions, called \textit{s-propositions} (\textit{s} for ``senses''), of the following form:
    \begin{equation}\label{eq:sProposition}
        S \senses F \withAccuracies M
    \end{equation}
    for an action $S$, a fluent $F$ and some matrix $M$ representing the accuracy of $A$ when sensing $F$. Given their role within EPEC, we would need s-propositions to model sensors in our domain. However, s-propositions add a layer of complexity that in this work we would like to avoid due to the necessity of taking decisions at runtime. In this section, we informally show a possible translation procedure s-propositions into pairs of a c-proposition and an o-proposition. This aims to show that one can ``improperly'' use c-propositions and o-propositions (that are typically used for effectors) as a means to also model epistemic aspects of a domain. Whether a sound and complete translation is available remains however a subject for future work.

    To illustrate, consider a simple domain description in which an agent (imperfectly) tests twice for $\Flu$:
    \begin{gather}
        \Flu \takesValues \{ \True, \False \}\\
        \initiallyOneOf \{ ( \{ \Flu \}, 0.7 ), ( \{ \neg \Flu \}, 0.3 ) \}\\
        \Test \senses \Flu \withAccuracies \begin{pmatrix}0.8 & 0.2\\0.4 & 0.6\end{pmatrix}\label{eq:exampleSprop}\\
        \Test \performedAt 0\label{eq:examplePProp1}\\
        \Test \performedAt 1\label{eq:examplePProp2}
    \end{gather}
    which means that the $\Test$ has the effect of producing knowledge about whether the agent has $\Flu$ with associated confusion matrix $\begin{pmatrix} 0.8 & 0.2\\0.4 & 0.6\end{pmatrix}$, where $0.8$ is the true positives rate and $0.6$ is the true negatives rate. The domain description above entails the following \textit{b-proposition} (\textit{b} for ``believes''):
    \begin{align}
        \at 2 &\believes \ifor{\Flu}{0} \withProbs\label{eq:beginningOfBprop}\\ \{&(\langle\{\Test,((\Test,\Flu),\false)\}\atsign 0,\{\Test,((\Test,\Flu),\false)\}\atsign 1\rangle,0.136,0.206),\label{eq:testnegativetwice}\\
        &(\langle\{\Test,((\Test,\Flu),\false)\}\atsign 0,\{\Test,((\Test,\Flu),\true)\}\atsign 1\rangle,0.184,0.609),\label{eq:testnegativepositive}\\
        &(\langle\{\Test,((\Test,\Flu),\true)\}\atsign 0,\{\Test,((\Test,\Flu),\false)\}\atsign 1\rangle,0.184,0.609),\label{eq:testpositivenegative}\\
        &(\langle\{\Test,((\Test,\Flu),\true)\}\atsign 0,\{\Test,((\Test,\Flu),\true)\}\atsign 1\rangle,0.496,0.903)\}\label{eq:testpositivetwice}
    \end{align}
    
    This proposition may be interpreted as saying that an agent sitting at instant $2$, having tested for $\Flu$ twice, is in one of the four following knowledge states:
    \begin{itemize}
        \item The test came up negative twice (as in the case of \eqref{eq:testnegativetwice}). Given the initial knowledge about the distribution of $\Flu$, this is likely to happen with probability $0.136$. In this case, the agent will believe that the probability of $\ifor{\Flu}{0}$ decreased from the initial value $0.7$ to $0.206$.
        \item The test came up negative at instant $0$ and positive at instant $1$ (as in the case of \eqref{eq:testnegativepositive}). This is likely to happen with probability $0.184$, and in this case the agent will believe that the probability of $\ifor{\Flu}{0}$ is $0.608$.
        \item The test came up positive at instant $0$ and negative at instant $1$ (as in the case of \eqref{eq:testpositivenegative}). This is likely to happen with probability $0.184$, and in this case the agent will believe that the probability of $\ifor{\Flu}{0}$ is $0.608$.
        \item The test came up positive twice (as inthe case of \eqref{eq:testpositivetwice}). This is likely to happen with probability $0.496$, and in this case the agent will believe that the probability of $\ifor{\Flu}{0}$ is $0.903$.
    \end{itemize}
    
    This syntax shows the epistemic core of EPEC, which allows one to reasoning about \textit{past, present and future}: for example, note that in \eqref{eq:beginningOfBprop} we reason about the perspective of an agent at instant $2$ who is reasoning about whether it \textit{had} $\Flu$ at instant $0$ in the past. This powerful syntax comes at expenses of computational efficiency, and therefore in this work we adopt a number of simplifications due to the fact that we only need to reason about the present. The first of such simplifications is that we drop s-propositions and simulate them via o-propositions and c-propositions, in a way that we now aim to make clearer.
    
    Assume that the agent modeled in the domain description above tests positive twice. Then, we may note that substituting the s-proposition \eqref{eq:exampleSprop} and the two p-propositions \eqref{eq:examplePProp1} and \eqref{eq:examplePProp2} with
    \begin{gather}
        \Test \causes \{ ( \{ \Flu \}, 1 ) \}\\
        \Test \occursAt 0 \withProb 0.412\\
        \Test \occursAt 1 \withProb 0.452
    \end{gather}
    yields
    \begin{equation}
        \mathcal{D} \mmodels \ifor{\Flu}{2} \holdsWithProb 0.903
    \end{equation}
    which matches with the result in \eqref{eq:testpositivetwice}, and therefore we may say that \textit{simulates} it. Although a formal characterization of the translation from s-proposition to c-propositions and o-propositions is beyond the scope of this work, note that an s-proposition for a boolean fluent $F$ of the form
    \begin{equation}
        S \senses F \withAccuracies \begin{pmatrix}a & 1-a\\b & 1-b\end{pmatrix}
    \end{equation}
    is simulated either by a c-proposition
    \begin{equation}
        S^+ \causesOneOf \{ ( \{ F \}, 1 ) \}
    \end{equation}
    and an o-proposition
    \begin{equation}
        S^+ \occursAt I \withProb \frac{p(a-b)}{p(a-b)+b}
    \end{equation}
    if $S$ is performed at $I$ and produces a positive result, or by a c-proposition
    \begin{equation}
        S^- \causesOneOf \{ ( \{ \neg F \}, 1 ) \}
    \end{equation}
    and an o-proposition
    \begin{equation}
        S^- \occursAt I \withProb \frac{(p-1)(a-b)}{p(a-b)+b-1}
    \end{equation}
    if $S$ is performed at $I$ and produces a negative result, where $p$ is the probability such that
    \[
        \mathcal{D}\mmodels \ifor{F}{I} \withProb p
    \]
    in the domain $\mathcal{D}$ under consideration. This informally shows that it is possible to meaningfully interpret c-propositions and o-propositions as epistemic actions in place of s-propositions, and justifies their employment to perform sensing.
    
\section{A runtime adaptation of EPEC}\label{sec:ARuntimeEPEC}
    The subset of EPEC introduced in Section \ref{sec:OverviewOfTheLanguage} is sufficient to handle domains that require limited epistemic functionalities. For this special class of domains, one can use the non-epistemic framework to \textit{approximate} the behavior of EPEC and efficiently compute probabilities of queries at runtime, at the expense of the ability to reason about the past. These assumptions are not restrictive for our use-case, as we only need to reason about the present state of the world and possibly react at runtime. In this section, we provide details about an implementation, dubbed \textit{PEC-RUNTIME}, that exploits these restrictions to provide a fast reasoning system that overcomes the limitations of other implementations of EPEC in terms of computation time. PEC-RUNTIME is publicly available\footnote{ \url{https://gitlab.com/fdasaro/pec-runtime}}.
    
    PEC-RUNTIME is based on \textit{PEC-ASP}, an Answer Set Programming implementation of the non-epistemic fragment of EPEC that was presented in \cite{dasaro2017lpnmr} and based on ASP grounder and solver clingo \cite{gebser2014clingo}. PEC-ASP and PEC-RUNTIME share the same syntax and domain-independent part of the implementation. In addition, PEC-RUNTIME exploits clingo's integration with Python to control the grounding and solving process.
    
    One of the main problems of PEC-ASP is that it is an \textit{exact} inference mechanism, which needs to enumerate all the well-behaved worlds that satisfy a given query. The number of worlds typically grows exponentially with the size of the narrative. As an illustration, consider the simple domain consisting of one fluent $F$, one action $A$, set of instants $\{ 0, 1, \dots, N-1 \}$ for some $N>0$, the i-proposition
    \begin{equation}
        \initiallyOneOf \{ (F,1) \}    
    \end{equation}
    
    and $N$ o-propositions
    \begin{equation}
        A \performedAt I \withProb 0.5
    \end{equation}
    one for each $I$ in the set of instants. This simple domain description has well-behaved worlds of the following form:
    \begin{center}
    		\begin{tikzpicture}[scale=0.9]
    		\draw[<-] (10,0) -- (-2,0) node[anchor=east] {};
    		\draw (0,-0.1) -- (0,0.1) node[anchor=south,text width=2.5cm,align=center] {\small$\{ F,$ $\pm A\}$};
    		\draw (3.5,-0.1) -- (3.5,0.1) node[anchor=south,text width=2.5cm,align=center] {\small$\{ F,$ $\pm A\}$};
    		\draw (8,-0.1) -- (8,0.1) node[anchor=south,text width=2.5cm,align=center] {\small$\{ F,$ $\pm A\}$};
    		\node at (0,-0.3) {{\small {$0$}}};
    		\node at (3.5,-0.3) {{\small {$1$}}};
    		\node at (5.75,-0.3) {{\small {\dots}}};
    		\node at (5.75,0.3) {{\small {\dots}}};
    		\node at (8,-0.3) {{\small {$N-1$}}};
    		\node at (10,-0.3) {{\small {$I$}}};
    		\end{tikzpicture}
    \end{center}
    where $\pm A$ is either $A$ or its negation $\neg A$. Clearly, there are $2^N$ such worlds. Given that $F$ is always true in all of them, PEC-ASP needs to enumerate all $2^N$ well-behaved worlds even to answer a simple query such as $\ifor{F}{N-1}$. A strategy to tackle this problem is to sample a number $M \ll 2^N$ of well-behaved worlds and approximate the probability of the query. This is the approach adopted by \textit{PEC-ANGLICAN}, a probabilistic programming implementation of PEC\footnote{\url{https://github.com/dasaro/pec-anglican}}. However, this technique also suffers from scalability and precision issues when dealing with large narratives as we describe in Section \ref{section:scalability}, Experiment (a), and is mostly suited for offline tasks.
    
    The approach we adopt here relies on a form of \textit{progression} \cite{lin1997howtoprogress}. Given a domain description $\mathcal{D}$ and an instant $I$, knowledge about what happened before instant $I$ can be appropriately recompiled and mapped to a new domain description $\mathcal{D}_{\geq I}$ that does not include any narrative knowledge about instants $<I$ but agrees with the original domain $\mathcal{D}$ on all queries about instants $\geq I$. This requires \textit{exhaustively} querying $\mathcal{D}$ about the state of the environment at $I$, i.e., if $F_1, \dots, F_n$ are all the (boolean) fluents in the language, one must perform queries $\ifor{\pm F_1 \wedge \dots \wedge \pm F_n}{I}$ where $\pm F_i$ is either $F_i$ or its negation $\neg F_i$. These queries must be then recompiled into an appropriate i-proposition.

    We elaborate this procedure using our Scenario \ref{ex:RunningExample}. Consider the domain description $\mathcal{D}$ and instant $1$. Since this domain has only one fluent, exhaustively querying $\mathcal{D}$ at $1$ means finding the probabilities $P_1$ and $P_2$ such that
    \begin{gather}
        \mathcal{D} \mmodels \ifor{\Attention}{1} \holdsWithProb P_1\\
        \mathcal{D} \mmodels \ifor{\neg \Attention}{1} \holdsWithProb P_2
    \end{gather}
    
    These values are $P_1 = 0.325$ and $P_2 = 1 - P_1 = 0.675$ (as it can be calculated e.g. by using PEC-ASP) and can be recompiled into the following i-proposition:
    \begin{equation}\label{eq:NewIProposition1}
        \initiallyOneOf \{ ( \{ \Attention \},0.325 ), ( \{ \neg \Attention \},0.675 ) \}
    \end{equation}
    
    The domain description $\mathcal{D}_{\geq 1}$ consists of propositions (\ref{eq:NewIProposition1}), (\ref{eq:exampleVProp}), (\ref{eq:exampleCProp1}), (\ref{eq:exampleCProp2}), (\ref{eq:examplePProp}) and (\ref{eq:exampleOProp2}). Since the idea is that of discarding all knowledge about instants before $1$, $\mathcal{D}_{\geq 1}$ has corresponding domain language with set of instants $\{ 1,2 \}$ (i.e., instant $0$ was removed from the language). Domain descriptions $\mathcal{D}$ and $\mathcal{D}_{\geq I}$ agree on all queries with instants $\geq I$: for example, it is possible to show that $M_{\mathcal{D}_{\geq 1}} (\ifor{\Attention}{2}) = 0.158275$, and we have already calculated it is also the case that $M_{\mathcal{D}}(\ifor{\Attention}{2}) = 0.158275$. Since domain description $\mathcal{D}_{\geq 1}$ has fewer o-propositions than $\mathcal{D}$, it also has fewer well-behaved worlds (namely $5$). In turn this intuitively means that, if a query only contains instants $\geq 1$, it is computationally more convenient to query $ \mathcal{D}_{\geq 1} $ rather than $\mathcal{D}$.
    
    Repeating the process on $\mathcal{D}$ and instant $2$ produces the i-proposition
    \begin{equation}\label{eq:NewIProposition2}
        \initiallyOneOf \{ ( \{ \Attention \},0.158275 ), ( \{ \neg \Attention \},0.841725 ) \}
    \end{equation}
    and the domain description $\mathcal{D}_{\geq 2}$ consisting of propositions (\ref{eq:NewIProposition2}), (\ref{eq:exampleVProp}), (\ref{eq:exampleCProp1}), (\ref{eq:exampleCProp2}) and (\ref{eq:examplePProp}). This new domain description has only $2$ well-behaved worlds. Clearly, $ M_{\mathcal{D}_{\geq 2}} (\ifor{\Attention}{2}) = 0.158275 $ as this is encoded directly in the i-proposition, and this also equals $M_{\mathcal{D}} (\ifor{\Attention}{2})$ and $M_{\mathcal{D}_{\geq 1}} (\ifor{\Attention}{2})$ as we have already calculated.
    
    Note that the domain description $\mathcal{D}_{\geq 2}$ trivializes the task of deciding whether Proposition (\ref{eq:examplePProp}) fires or not. In fact, it suffices to check whether its epistemic precondition is \textit{satisfied} or not by the i-proposition in $\mathcal{D}_{\geq I}$. In this case, the epistemic precondition requires $\Attention$ to be in the interval $[0,0.1]$, which is not the case as it can be immediately seen by looking at the i-proposition (\ref{eq:NewIProposition2}) and considering that $0.158275 \notin [0,0.1]$.
    
    As we show below, we can turn this intuition into an efficient procedure for updating a domain description as new events are received and reason about this smaller domain, instead of augmenting it with new propositions and reason about the full augmented domain. The pseudo code of our PEC-RUNTIME procedure is as follows:

    \begin{algorithm}
    \SetAlgoLined
     I = 0\;
     \While{session is not over}{
      $\currentIProp \leftarrow \textit{constructIProp} (\textit{exhaustivelyQuery} (\mathcal{D},I))$\;
      $\mathcal{D} \leftarrow \mathcal{D} \setminus \iprop(\mathcal{D}) \setminus \narr(\mathcal{D}) \cup \{ \currentIProp \}$\;
      \For{p-proposition $p$ in $\mathcal{D}$}{
       \If{$p$ is satisfied in $\iprop(\mathcal{D})$ }{
        Execute $p$'s postconditions\;
      }}
      $\currentNarrative \leftarrow \generateEventsFrom(C,I)$\; 
      $\mathcal{D} \leftarrow \mathcal{D} \cup \currentNarrative$\;
      $\outputNarrative \leftarrow \outputNarrative \cup \currentNarrative$\;
      $I \leftarrow I+1$\;
     }
     \Return{\outputNarrative}
     \caption{\textit{PEC-RUNTIME}$($\textit{Domain Description} $\mathcal{D},$ \textit{Classifiers } $C)$}
    \end{algorithm}
    
    The correctness of this algorithm is guaranteed by the following proposition. 
    
    \begin{proposition}
        Let $\mathcal{D}$ be any domain description such that $\mathcal{D} = \mathcal{D}_{\leq 0}$.
        Then, $M_{\mathcal{D}_{>0}}(\varphi) = M_{\mathcal{D}} (\varphi)$ for any i-formula $\varphi$ having instants $>0$.
    \end{proposition}

    \begin{proof}
        See Appendix B
    \end{proof}

\section{Sensors and Classifiers}\label{sec:Architecture}

    \begin{figure}
    \begin{center}
    \includegraphics[width=7cm]{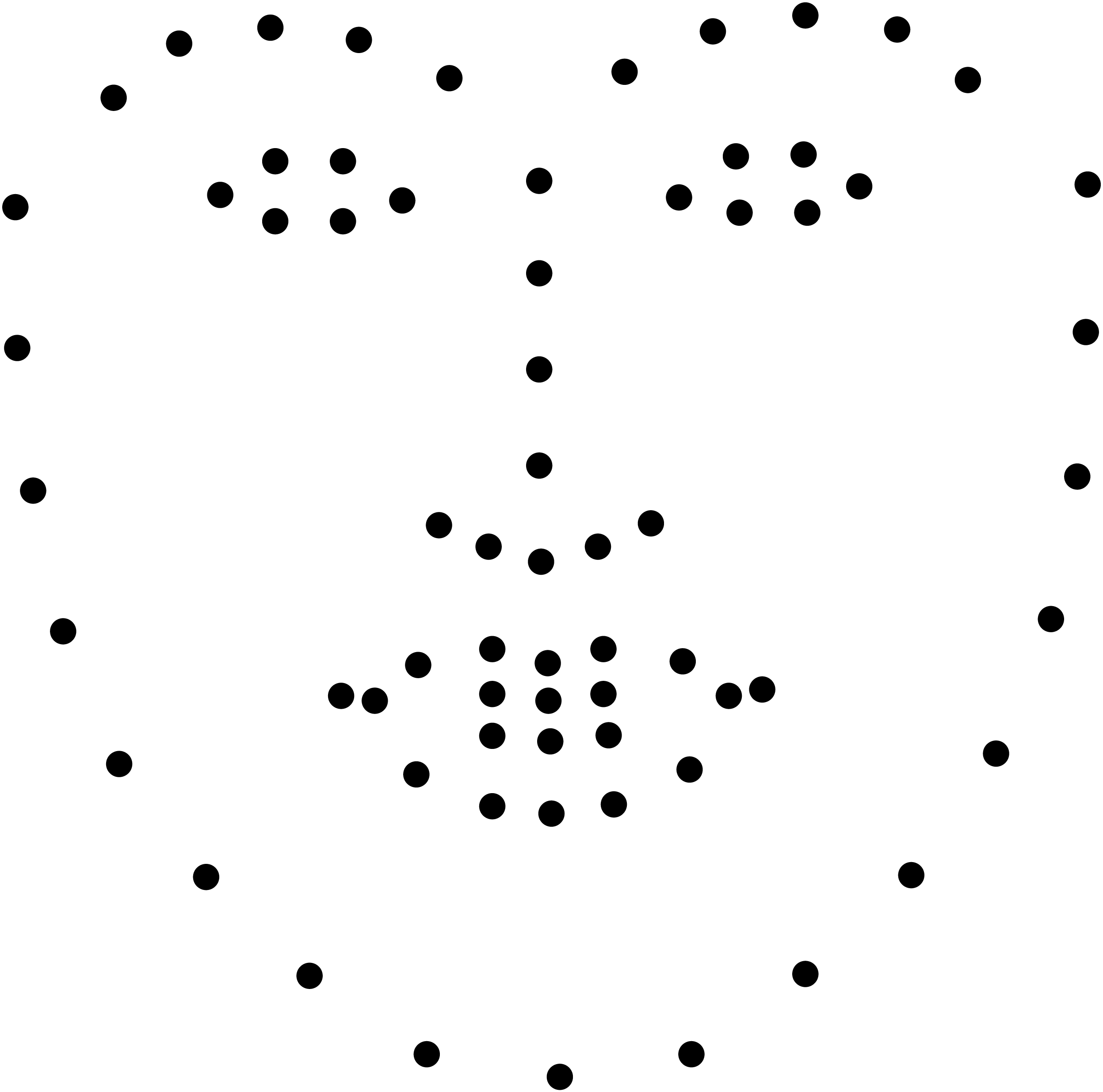}
    \end{center}
    \caption{Facial landmarks detected by the Head Pose Recognition module.} \label{fig:facial_landmarks}
    \end{figure}
    
    In this section, we briefly overview the layers of sensors and classifiers we currently use to detect user activity that is relevant to the rehabilitation task.
    
    At the present stage, our main source of information is a 2D-webcam placed in front of the children performing the rehabilitation task. The camera stream is mainly responsible for the evaluation of the current pose of head and shoulders of the child, and to estimate his/her emotional response to the interaction. Its output is processed by three modules:
    
        \textbf{Head Pose Recognition Module}: We use the library \textit{Head Pose Estimation}\footnote{\url{https://github.com/yinguobing/head-pose-estimation}} to detect the facial landmarks (see Figure \ref{fig:facial_landmarks}). A box containing the face and its surrounding area is selected, resized, and normalized in order to use it as input for the facial-landmarks detector, which is based on a CNN architecture. The facial-landmarks detection step is carried out by a custom trained facial landmark detector based on TensorFlow and trained on the iBUG datasets\footnote{\url{https://ibug.doc.ic.ac.uk/}}. It outputs $68$ facial landmarks (2D points) as in Figure \ref{fig:facial_landmarks}. In turn, these landmarks are used to estimate the direction of gaze and quality of head posture \cite{MALEK2021406}.
        
        \textbf{Shoulders Alignment Module}: We use \textit{PoseNet} \cite{papandreou2018personlab} to extract the skeleton joints coordinates from the camera frames. Then we compare these coordinates to the horizontal plane. We use this estimation to determine whether the child is assuming a correct posture while performing the rehabilitation.
        
        \textbf{Emotion Recognition Module}: This module estimates the emotional response of the child, and outputs a \textit{valence} \cite{Rescigno2020}. This value represents how good - or how bad - the emotion associated with the event or the situation was, and ranges from $-1$ to $+1$. To this end we use the \textit{AlexNet} model \cite{krizhevsky2012imagenet} trained on the \textit{Affectnet} Dataset \cite{mollahosseini2017affectnet}.

    In future extensions of this work we intend to combine data gathered with additional sensory sources. Specifically, we will employ pillows equipped with pressure sensors, mounted on the seat-back of the chair, to complement the evaluation of the Shoulders Alignment Module and better evaluate whether the child is sitting correctly. We also plan on complementing the Head Pose and Emotion Recognition Modules using an Eye Tracker and EEG respectively.
    
\section{Examples and discussion}\label{sec:Examples}
    
    In this section we demonstrate our approach by means of examples and experimental results. Note that Experiment (a) was run on a Mid-2010 Apple MacBook Core 2 Duo $2.4$ GHz with $8$ GB of RAM, while Experiments (b), (c), (d), (e) and (f) were run on an Apple Macbook Pro 2020 M1 with $16$ GB of RAM.
    
    \subsection{Scalability}\label{section:scalability}
    
    \textbf{Experiment (a).} To demonstrate that PEC-RUNTIME scales favourably compared with PEC-ASP and PEC-ANGLICAN we tested the three implementations on a simple domain description in which an action repeatedly occurs and causes the probability of a fluent to slowly decay:
    \begin{gather*}
        F \takesValues \{ \True, \False \}\\
        \initiallyOneOf \{ (\{ F \}, 1 ) \}\\
        A \causesOneOf \{ (\{ \neg F \},0.2), (\emptyset,0.8) \}\\
        \forall I,\ A \performedAt I \withProb 0.5
    \end{gather*}
    where the set of instants is $\{ 0, 1, \dots, 15 \}$. The probability of $\neg F$ as calculated by the three frameworks is shown in Figure \ref{fig:decayExactProbability}. Table \ref{tab:scalability} summarizes the performances of PEC-RUNTIME, PEC-ASP and PEC-ANGLICAN.
    
    As it can be seen from these results, PEC-RUNTIME outperforms both PEC-ASP and PEC-ANGLICAN on all queries. In terms of performance, PEC-ANGLICAN is close to PEC-RUNTIME when only $100$ well-behaved worlds are sampled. However, sampling $100$ worlds causes precision issues as it is clear from Figure \ref{fig:decayExactProbability}.
    
    \begin{table}[]
    \centering
    \begin{tabular}{cccccc}
    \hline\hline
    Instant $I$ & PEC-ASP & PEC-ANG (100) & PEC-ANG (1000) & PEC-ANG (10000) & PEC-RUNTIME \\
    \hline
    0           & 0.01    & 1.55          & 4.52           & 29.29           & 0.07\\
    1           & 1260.24 & 1.58          & 3.23           & 29.02           & 0.19\\
    2           & 2142.81 & 1.64          & 4.87           & 31.54           & 0.18\\
    3           & 2557.31 & 1.58          & 5.46           & 33.04           & 0.17\\
    4           & 2708.37 & 1.61          & 3.22           & 28.66           & 0.17\\
    5           & 2911.29 & 1.61          & 3.69           & 32.92           & 0.22\\
    6           & 3086.32 & 1.79          & 3.58           & 35.62           & 0.13\\
    7           & 3147.04 & 1.56          & 3.70           & 26.98           & 0.22\\
    8           & 3208.48 & 1.55          & 5.18           & 33.00           & 0.14\\
    9           & 3251.10 & 1.53          & 3.33           & 32.79           & 0.19\\
    10          & 3298.81 & 1.53          & 5.41           & 32.95           & 0.18\\
    11          & 3316.88 & 1.60          & 4.66           & 28.72           & 0.13\\
    12          & 3311.27 & 1.57          & 3.88           & 27.99           & 0.21\\
    13          & 3338.23 & 1.55          & 1.50           & 27.62           & 0.18\\
    14          & 3440.68 & 1.56          & 4.05           & 23.97           & 0.15\\
    15          & 3368.67 & 1.80          & 4.65           & 24.85           & 0.13\\
    \hline
    Average:         & 2771.72 & 1.60          & 4.06           & 29.93           & 0.17\\
    \hline\hline
    \end{tabular}
    \caption{Time (in seconds) to execute the query $\ifor{\neg F}{I}$ in the example discussed in Section \ref{section:scalability}. The numbers in bracket in the case of PEC-ANGLICAN refer to the number of sampled well-behaved worlds used to approximate the result of the query. For every implementation, reported times include grounding and processing of the domain description.}\label{tab:scalability}
    \end{table}
    
    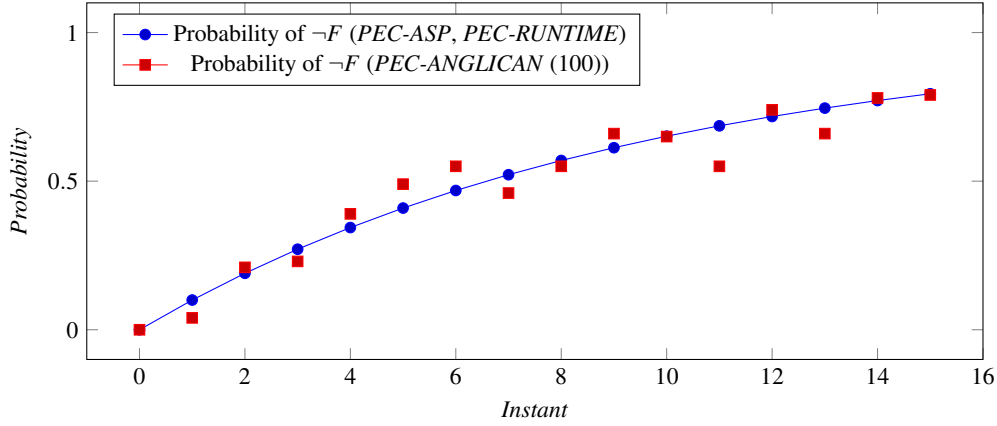
\begin{figure}
        \centering
        \begin{tikzpicture}
            \begin{axis}[
                width=\textwidth,
                height=20em,
                legend pos=north west,
                ylabel=\textit{Probability},
                xlabel=\textit{Instant},
                xmin=-1, xmax=16,
                ymin=-0.1, ymax=1.1
            ]
            \addplot table [x=instant,y=probn,col sep=comma]{Data/Decay/theoretical.csv};
            \addlegendentry{Probability of $\neg F$ (\textit{PEC-ASP}, \textit{PEC-RUNTIME})}
            
            \addplot table [x=inst,y=probn,col sep=comma,only marks]{Data/Decay/ANG/100.csv};
            \addlegendentry{Probability of $\neg F$ (\textit{PEC-ANGLICAN} ($100$))}
            
            \end{axis}
        \end{tikzpicture}
        \caption{Probability of $F$ as a function of time for the example discussed in Section \ref{section:scalability}, Experiment (a).}
        \label{fig:decayExactProbability}
    \end{figure}
    
    
    \textbf{Experiment (b).} In this experiment we consider the effect of varying the number of actions in the domain description. To isolate the effect of actions from that of other side-effects we consider domain descriptions of the simplest form possibile:
    \begin{gather*}
    F \takesValues \{ \top, \bot \}\\
    \initiallyOneOf \{ ( \{ F \}, 1 ) \}\\
    \forall I,\ A_1 \performedAt I \withProb 0.5\\
    \vdots\\
    \forall I,\ A_n \performedAt I \withProb 0.5
    \end{gather*}
    where $\mathcal{I} = \{ 1, \dots, 15 \}$ and we consider different values of $n$ ranging from $0$ to $15$. We evaluate the computation time for query $\ifor{F}{I}$ at all instants. Results are plotted in Figure \ref{fig:scalabilityactionandfluents} and confirm the intuition that, at an instant $I$, computation time scales exponentially with the number of action occurrences at that instant.

    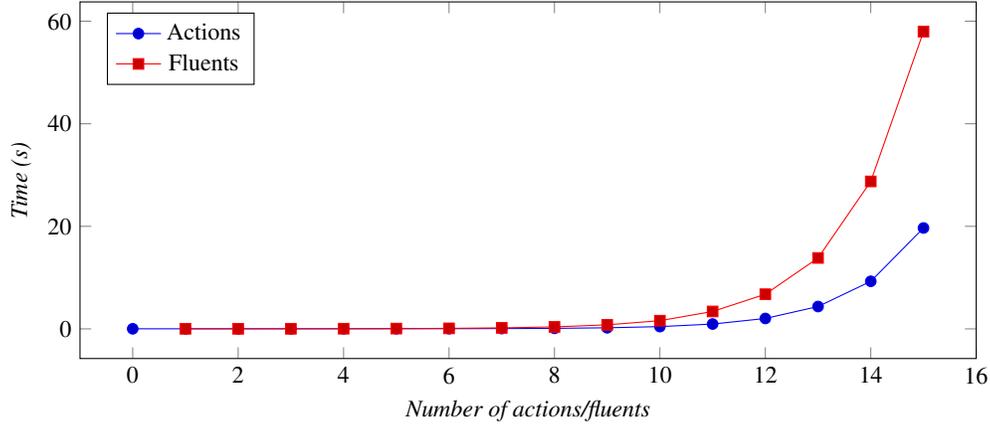
\begin{figure}
        \centering
            \begin{tikzpicture}
                \begin{axis}[
                    width=\textwidth,
                    height=20em,
                    legend pos=north west,
                    ylabel=\textit{Time (s)},
                    xlabel=\textit{Number of actions/fluents},
                    xmin=-1, xmax=16
                ]
                \addplot table [x=actions,y=average,col sep=comma]{Data/Scalability/actions.csv};
                \addlegendentry{Actions}
                
                \addplot table [x=fluents,y=average,col sep=comma]{Data/Scalability/fluents.csv};
                \addlegendentry{Fluents}
                \end{axis}
            \end{tikzpicture}
        \caption{Time (in seconds) to query the domain in Section
5.1, Experiments (b) and (c), expressed as a function of the number of actions and fluents. The results show averages over $15$ runs -- however, standard error was not plotted as it it significantly small ($<0.05$ at all data points).}\label{fig:scalabilityactionandfluents}
    \end{figure}
    
    \textbf{Experiment (c).} To empirically evaluate the effect of the number of fluents in the domain, we performed a similar experiment with domain descriptions of the following forms:
    \begin{gather*}
    F_1 \takesValues \{ \top, \bot \}\\
    \vdots\\
    F_n \takesValues \{ \top, \bot \}\\
    \initiallyOneOf \{ ( \{ F_1, \dots, F_n \}, 1 ) \}
    \end{gather*}
    where $\mathcal{I}=\{ 1, \dots, 15 \}$ and $n$ ranges from $1$ to $15$. Results are shown in Figure \ref{fig:scalabilityactionandfluents}, and confirm the intuition that computational time scales exponentially as a function of the number of fluents in the domain.
    
\textbf{Experiment (d).} This experiment combines (b) and (c) in that it shows how computation time scales with both fluents and actions. The considered domain descriptions are:
    \begin{gather*}
    F_1 \takesValues \{ \top, \bot \}\\
    \vdots\\
    F_n \takesValues \{ \top, \bot \}\\
    \initiallyOneOf \{ ( \{ F_1,\dots,F_n \}, 1 ) \}\\
    \forall I,\ A_1 \performedAt I \withProb 0.5\\
    \vdots\\
    \forall I,\ A_m \performedAt I \withProb 0.5
    \end{gather*}
    where $\mathcal{I}=\{ 1,\dots,15 \}$, the number of fluents varies from $1$ to $15$ and the number of actions varies from $0$ to $15$. Result are shown in Figure \ref{fig:scalabilitysummary} and extend those of Figure \ref{fig:scalabilityactionandfluents}.
    
    \begin{figure}
        \centering
            \includegraphics[]{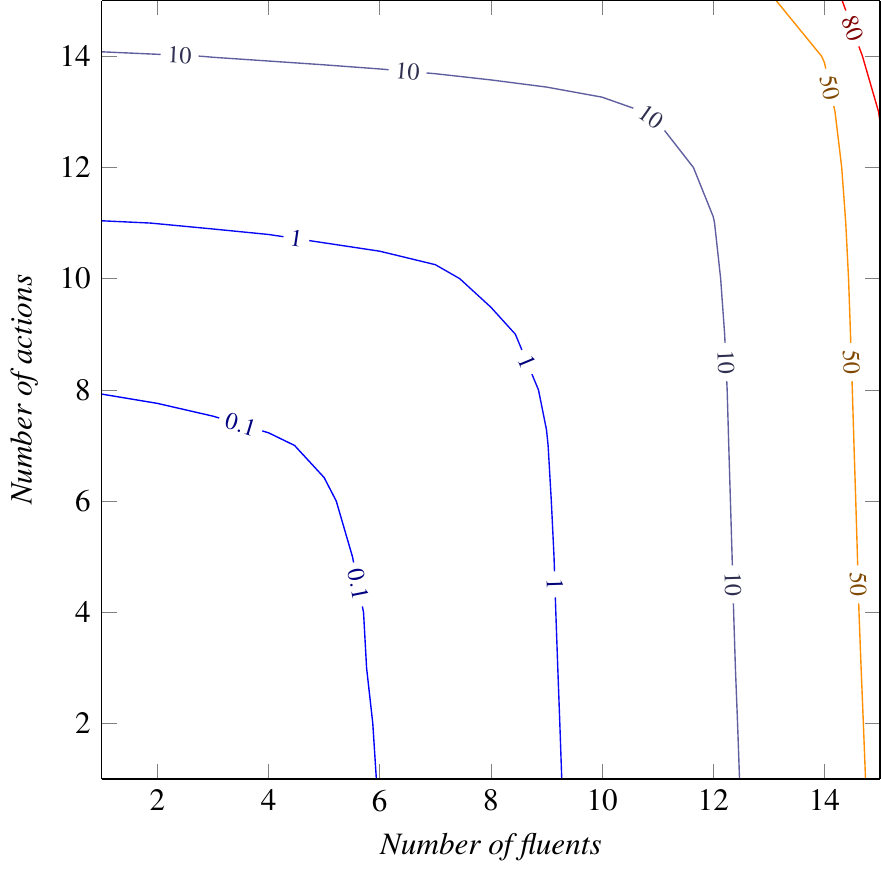}
        \caption{Contour plot showing time (in seconds) to query the domain in Section
5.1, Experiment (d), expressed as a function of the number of actions and fluents.}
        \label{fig:scalabilitysummary}
    \end{figure}
    
    \textbf{Experiment (e).} In this experiment we empirically evaluate the effect of initial conditions, and show that computation time scales linearly with their number. To this aim, we consider the following domain description:
    \begin{gather*}
    F_1 \takesValues \{ \top, \bot \}\\
    \vdots\\
    F_{10} \takesValues \{ \top, \bot \}\\
    \initiallyOneOf \{ ( \{ F_1, \dots, F_{10} \}, 1/n ), \dots, ( \{ \neg F_1, \dots, F_{10} \}, 1/n ), ( \{ \neg F_1, \dots, \neg F_{10} \}, 1/n ) \}
    \end{gather*}
    where $\mathcal{I}=\{ 1, \dots, 15 \}$ and $n$ ranges from $1$ to $2^{10}$, i.e., the maximum number of possible initial conditions with $10$ fluents. Note that $n$ is also the number of well-behaved worlds considered by PEC-RUNTIME at each progression step. Results are shown in Figure \ref{fig:scalabilityiconds} and confirm the intuitive hypothesis that computational time scales linearly with the number of initial conditions/well-behaved worlds.
    
    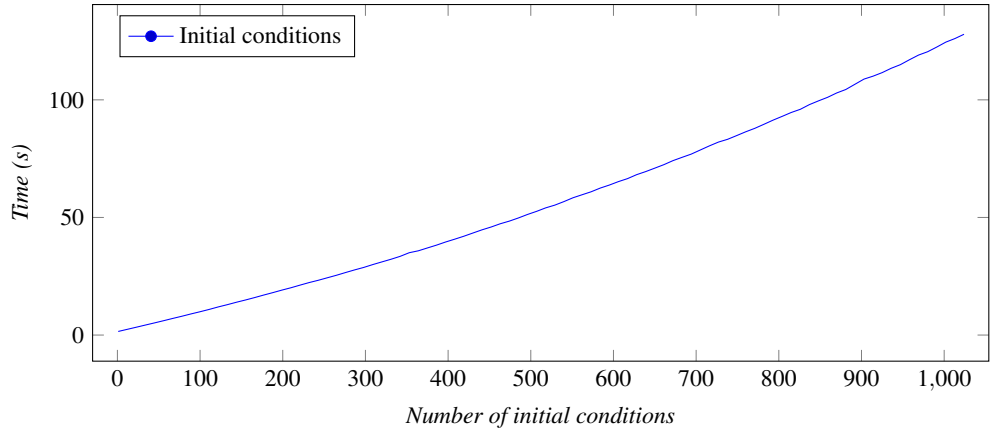
\begin{figure}
        \centering
            \begin{tikzpicture}
                \begin{axis}[
                    width=\textwidth,
                    height=20em,
                    legend pos=north west,
                    xmin = -30, xmax = 1054,
                    ylabel=\textit{Time (s)},
                    xlabel=\textit{Number of initial conditions},
                ]
                \addplot table [x=iconds,y=time,col sep=comma,mark=none]{Data/Scalability/iconds.csv};
                \addlegendentry{Initial conditions}

                \end{axis}
            \end{tikzpicture}
        \caption{Time (in seconds) to query the domain in Section
5.1, Experiment (e), expressed as a function of the number of initial conditions.}
        \label{fig:scalabilityiconds}
    \end{figure}
    
    \textbf{Experiment (f).} This experiment aims to study how PEC-RUNTIME behaves in a more realistic domain where every sensor has a fixed probability $p$ of producing a reading at every instant. We consider the following domain descriptions:
    \begin{gather*}
    F_1 \takesValues \{ \top, \bot \}\\
    F_2 \takesValues \{ \top, \bot \}\\
    F_3 \takesValues \{ \top, \bot \}\\
    F_4 \takesValues \{ \top, \bot \}\\
    F_5 \takesValues \{ \top, \bot \}\\
    \initiallyOneOf \{ ( \{ \neg F_1, \neg F_2, \neg F_3, \neg F_4, \neg F_5 \}, 1 ) \}\\
    \neg A_1 \wedge \neg A_2\wedge \neg A_3 \wedge \neg A_4 \wedge A_5  \causesOneOf \{ (\{ \neg F_1, \neg F_2, \neg F_3, \neg F_4, F_5 \},4/5), (\emptyset,1/5) \}\\
    \vdots\\
    A_1 \wedge A_2\wedge A_3 \wedge A_4 \wedge \neg A_5  \causesOneOf \{ (\{ F_1, F_2, F_3, F_4, \neg F_5 \},4/5), (\emptyset,1/5) \}\\
    A_1 \wedge A_2\wedge A_3 \wedge A_4 \wedge A_5  \causesOneOf \{ (\{ F_1, F_2, F_3, F_4, F_5 \},4/5), (\emptyset,1/5) \}\\
    \end{gather*}
    and, for every instant $I$ and action $A$, the proposition
    \[
        A_i \performedAt I \withProb 0.5
    \]
    has a some fixed probability $p$ of being included in the domain description, where $p$ varies in the set $\{ 0.1, 0.2, \dots, 1 \}$. Results are shown in Figure \ref{fig:scalabilitydensity}.
    
    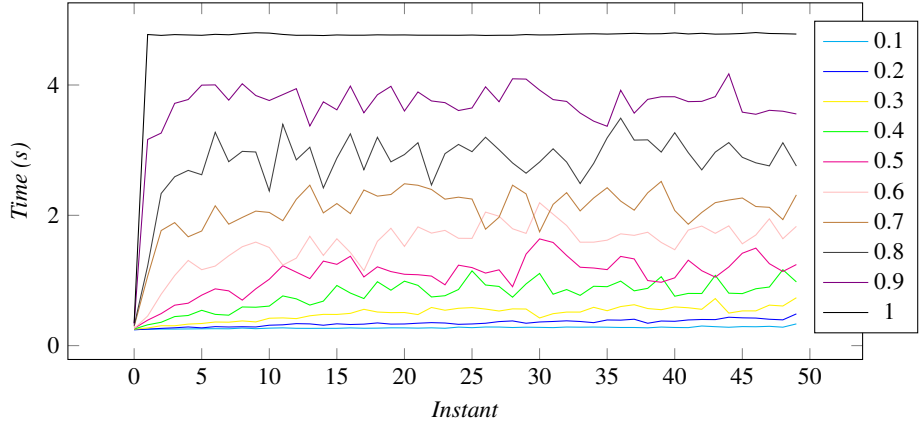
\begin{figure}
        \centering
            \begin{tikzpicture}
                \begin{axis}[
                    width=0.9\textwidth,
                    height=20em,
                    legend style={ at={(1.07,0.95)} },
                    ylabel=\textit{Time (s)},
                    xlabel=\textit{Instant},
                ]
                \addplot[color=cyan] table [x=inst,y=0.1,col sep=comma,mark=none]{Data/Scalability/density.csv};
                \addlegendentry{$0.1$}
                \addplot[color=blue] table [x=inst,y=0.2,col sep=comma,mark=none]{Data/Scalability/density.csv};
                \addlegendentry{$0.2$}
                \addplot[color=yellow] table [x=inst,y=0.3,col sep=comma,mark=none]{Data/Scalability/density.csv};
                \addlegendentry{$0.3$}
                \addplot[color=green] table [x=inst,y=0.4,col sep=comma,mark=none]{Data/Scalability/density.csv};
                \addlegendentry{$0.4$}
                \addplot[color=magenta] table [x=inst,y=0.5,col sep=comma,mark=none]{Data/Scalability/density.csv};
                \addlegendentry{$0.5$}
                \addplot[color=pink] table [x=inst,y=0.6,col sep=comma,mark=none]{Data/Scalability/density.csv};
                \addlegendentry{$0.6$}
                \addplot[color=brown] table [x=inst,y=0.7,col sep=comma,mark=none]{Data/Scalability/density.csv};
                \addlegendentry{$0.7$}
                \addplot[color=darkgray] table [x=inst,y=0.8,col sep=comma,mark=none]{Data/Scalability/density.csv};
                \addlegendentry{$0.8$}
                \addplot[color=violet] table [x=inst,y=0.9,col sep=comma,mark=none]{Data/Scalability/density.csv};
                \addlegendentry{$0.9$}
                \addplot[color=black] table [solid,x=inst,y=1,col sep=comma,mark=none]{Data/Scalability/density.csv};
                \addlegendentry{$1$}

                \end{axis}
            \end{tikzpicture}
        \caption{Time (in seconds) to query the example discussed in Section
5.1, Experiment (f). The results show averages over $30$ runs. }
        \label{fig:scalabilitydensity}
    \end{figure}
    
    \subsection{Effect of Noise}
    One of the characteristics of EPEC (and its dialects) that makes it highly suitable as an interface with machine learning algorithms is that it supports events annotated with probabilities. If a classifier provides a confidence score for its output, it is possible to feed such score into EPEC in the form of a probability. This may be useful to account for possible artifacts and flickering of a given classifier. For instance, consider a classifier $C$ that alternatively produces readings $\true$ and $\false$ for some characteristic $F$. This may be represented in EPEC as the following stream of events:
    \begin{gather*}
        \Ctrue \occursAt 0 \withProb P_{true}\\
        \Cfalse \occursAt 1 \withProb P_{false}\\
        \Ctrue \occursAt 2 \withProb P_{true}\\
        \Cfalse \occursAt 3 \withProb P_{false}\\
        \vdots\\
        \Ctrue \occursAt 18 \withProb P_{true}\\
        \Cfalse \occursAt 19 \withProb P_{false}
    \end{gather*}
    where $\Ctrue$ (resp. $\Cfalse$) is an event that causes the value of fluent $F$ to be $\True$ (resp. $\False$), i.e.:
    \begin{gather*}
        \Ctrue \causesOneOf \{ (\{ F \}, 1) \}\\
        \Cfalse \causesOneOf \{ (\{ \neg F \}, 1) \}
    \end{gather*}
    and the truth value of $f$ is initially unknown, i.e.:
    \begin{gather*}
        \initiallyOneOf \{ (\{F\},0.5), (\{\neg F\},0.5) \}
    \end{gather*}
    We analyse the behavior of EPEC in different scenarios (corresponding to different values for $P_{true}$ and $P_{false}$):
    
    \textbf{Scenario (a).} The classifier $C$ produces $\true$ with high confidence and $\false$ with low confidence, e.g. $P_{true}=0.99$ and $P_{false}=0.01$. This domain description entails ``$\ifor{F}{20} \holdsWithProb P$'' where $P\approx 0.9899$, which makes intuitive sense as the $\false$ events have low confidence and are discarded as background noise.
    
    \textbf{Scenario (b).} The classifier $C$ produces $\true$ and $\false$ with a high degree of uncertainty. However, it assigns $\true$ a slightly higher degree of confidence, e.g. $P_{true}=0.501$ and $P_{false}=0.499$. This domain description entails ``$\ifor{F}{20} \holdsWithProb P$'' where $P\approx 0.653246$, which reflects that such a  sequence of events does not carry as much information as in Scenario (a), due to the the small difference between $P_{true}$ and $P_{false}$.
    
    \textbf{Scenario (c).} The classifier $C$ produces both $\true$ and $\false$ with low confidence, with $P_{true}\gg P_{false}$. For example, let $P_{true} = 0.49$ and $P_{false}=0.01$. This domain description entails ``$\ifor{F}{20} \holdsWithProb P$'' where $P\approx 0.979286$, which again makes intuitive sense as $\false$ readings are again discarded as background noise, with the repeated $\true$ readings making the probability of $F$ steadily grow throughout the timeline. 
    
    These scenarios show that EPEC is sensibly more accurate than any logical framework that is not capable of handling probabilities in the presence of noisy sensors. In fact, these frameworks would have to set a threshold, and only accept events with probabilities higher than that threshold. In our example, setting a threshold e.g. of $0.5$ would lead to significant errors especially in Scenarios (b) and (c). In Scenario (b), all $\Cfalse$ events would be discarded, implying that $F$ is detected to hold true at instant $20$ with certainty (compare this to the probability $0.653246$ assigned to the same query by EPEC). In Scenario (c), all events would be discarded and it would not be possible neither that $F$ holds at $20$ nor that it does not (compare again to the probability $0.979286$ assigned by EPEC).
    
    
    \subsection{An example from AVATEA}\label{sec:anExampleFromAvatea}
    As discussed in Section \ref{sec:Architecture}, a layer of classifiers directly injects events into PEC-RUNTIME, which merges them together and decides what action to take according to a predefined conditional plan. We show how PEC-RUNTIME accounts for these classifiers in the following example.
    
    We use the Head Pose Recognition module to decide whether the child is looking at the screen or not, and therefore this module can trigger an $\EyesNotFollowingTarget$ event (and possibly an associated probability). Similarly, the Shoulders Alignment Module triggers a $\BadPosture$ event if it detects that the child's shoulders are not correctly aligned. Finally, the Emotion Recognition Module triggers a $\LowAttention$ event when it thinks the child is experiencing negative valence. On the basis of these events, we aim to evaluate the probabilities of $\TaskCorrect$, i.e., that the child is performing the therapeutic task correctly, and $\Engagement$, i.e., that the child and engaged to the task. Consider the following narrative of events:

    \begin{gather*}
    \EyesNotFollowingTarget	\occursAt	0	\withProb	1\displaybreak[0]\\
    \BadPosture	\occursAt	0	\withProb	1\displaybreak[0]\\
    \LowAttention	\occursAt	0	\withProb	1\displaybreak[0]\\
    \EyesNotFollowingTarget	\occursAt	1	\withProb	1\displaybreak[0]\\
    \BadPosture	\occursAt	1	\withProb	1\displaybreak[0]\\
    \LowAttention	\occursAt	1	\withProb	1\displaybreak[0]\\
    \EyesNotFollowingTarget	\occursAt	2	\withProb	76/100\displaybreak[0]\\
    \LowAttention	\occursAt	2	\withProb	87/100\displaybreak[0]\\
    \EyesNotFollowingTarget	\occursAt	3	\withProb	1\displaybreak[0]\\
    \BadPosture	\occursAt	3	\withProb	1\displaybreak[0]\\
    \LowAttention	\occursAt	3	\withProb	1\displaybreak[0]\\
    \EyesNotFollowingTarget	\occursAt	7	\withProb	7/100\displaybreak[0]\\
    \EyesNotFollowingTarget	\occursAt	8	\withProb	89/100\displaybreak[0]\\
    \EyesNotFollowingTarget	\occursAt	9	\withProb	74/100\displaybreak[0]\\
    \EyesNotFollowingTarget	\occursAt	11	\withProb	1
    \end{gather*}
    
    These events are given a meaning by means of a domain description (see Appendix A for the full domain) that specify what bearing each event has on fluents $\TaskCorrect$ and $\Engagement$.
    
    Furthermore, consider a conditional plan of the following form:
    \begin{gather*}
        \forall I,\ \PlaySound \performedAt I \ifBelieves (\Attention,[0,0.3])\\
        \forall I,\ \LowerDifficultyLevel \performedAt I \ifBelieves (\TaskCorrect,[0,0.3])
    \end{gather*}
    which states that a sound must be played by the system whenever the $Attention$ of the child is low, and that the difficulty level must be lowered if the child does not manage to perform it correctly.
    
    \begin{figure}
        \centering
        \begin{tikzpicture}
            \begin{axis}[
                legend pos=south east,
                ylabel=\textit{Probability},
                xlabel=\textit{Instant},
                ymin=0, ymax=1.1
            ]
            \addplot table [color=red,x=Inst,y=Prob,col sep=comma]{Data/Example1/ASP/engagement.csv};
            \addplot table [color=blue,x=Inst,y=Prob,col sep=comma]{Data/Example1/ASP/taskCorrect.csv};
            \addplot[color=black, dashed, no marks] table [x=Inst,y=Prob,col sep=comma]{Data/Example1/threshold20.csv};
            \addlegendentry{Engagement}
            \addlegendentry{TaskCorrect}
            \addlegendentry{Threshold}
            \end{axis}
        \end{tikzpicture}
        \caption{$\Engagement$ and $\TaskCorrect$ as a function of time in the example from Section \ref{sec:anExampleFromAvatea}}
        \label{fig:avateaExample}
    \end{figure}
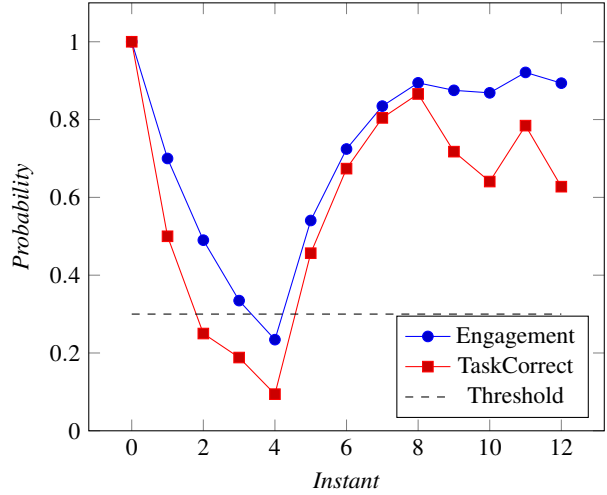
    
    PEC-RUNTIME works out the probabilities in Figure \ref{fig:avateaExample}, and triggers decisions when they fall below the given threshold. In particular, $\PlaySound$ is triggered at instant $4$, and $\LowerDifficultyLevel$ is triggered at instants $2$, $3$ and $4$. It is worth noting here that thresholds may induce the so-called \textit{rubberband effect}, where abrupt probability changes (due e.g. to noise) may lead to decisions being taken too frequently. Unfortunately, EPEC (and therefore also PEC-RUNTIME) does not allow to control this effect, therefore we implemented a bespoke semaphore in the underlying Python script that only allows decision to be taken every $20$ seconds.
    
    These decisions are forwarded to the game. At the end of each therapeutic session, this narrative is saved to a file so that it can be used to provide feedback to the therapists and re-consulted if needed. In this scenario, PEC-RUNTIME provides feedback such as ``The child performed the task correctly for a total of $10$ instants out of $13$ ($76.9\%$). The game difficulty was lowered at instants $2$, $3$ and $4$ as s/he was unable to perform the exercise. His attention was lowest at instant $4$.''. A graph similar to that in Figure \ref{fig:avateaExample} is also shown to the therapists, so that they can have a quick overview of the session.
    
\section{Related work}\label{sec:RelatedWork}
    To our knowledge, our architecture is one of the first that merges \textit{gamification} strategies for rehabilitation and \textit{probabilistic logic programming} techniques. In the following, we briefly survey these two fields, and motivate our design choices.
    
    \subsection{Gamification}
    \textit{Gamification} strategies consist in using game-like elements (e.g., points, rewards, performance graphs, etc.) in serious contexts such as ours. These have proven to be extremely successful to engage young children in diagnostic and therapeutic exercises, even before the advent of digital gaming.
    %
    While games to test cognitive capabilities \cite{Belpaeme2012} do not form a sharply defined class, games designed to test and improve motor skills are usually referred to as \emph{exergames}. The effects of exergames have been found to be generally positive \cite{Vernadakis2015}.
    Gamification systems are usually effective in engaging young users in playful activities, while adapting the current challenge according to level of user competence. Sessions are typically logged in order to provide detailed feedback to therapists. On the cognitive side, these adaptive systems have been designed to evaluate subjective well-being \cite{Wu2019} and phonological acquisition \cite{Origlia2017} among others. Adaptive exergames have been used e.g. in the context of children with spinal impairments \cite{Mulcahey2008}, and to test gross motor skills \cite{Huang2018}.
    
    (Deep) Machine Learning techniques are not advantageous in the case of exergames if they are used alone, as they must be trained on big amounts of data even when expert knowledge can be (relatively) easily extracted from experts and encoded in symbolic form (e.g., using logic programming), with the further advantage that symbolic knowledge can be used to automatically produce explanations and reports. As a centralized reasoning system, we use probabilistic logic programming techniques instead. These systems are a convenient option as they allow both for the representation of formal constraints needed to implement a clinically effective exercise, and for the statistical modeling of intrinsically noisy data sources.
    
    \subsection{Probabilistic Reasoning about Actions}
    Recently, logic-based techniques have been successfully applied to several fields of Artificial Intelligence, including among others event recognition from security cameras \cite{skarlatidis2015probabilistic,skarlatidis2015mlnec}, robot location estimation \cite{belle2018reasoning}, understanding of tenses \cite{van2008proper}, natural language processing \cite{nadkarni2011natural}, probabilistic diagnosis \cite{lee2018probabilistic} and intention recogntion \cite{acciaro2021predicting}. Given the importance of Machine Learning and Probability Theory in AI, these frameworks and languages have gradually started employing probabilistic semantics \cite{sato1995statistical} to incorporate and deal with uncertainty. This has given birth to the field of \textit{Probabilistic Logic Programming} \cite{riguzzi2018foundations}, and we are particularly concerned with Probabilistic Reasoning about Action frameworks as they can deal with agents interacting with some (partially known) environment. For example \cite{bacchus1999reasoning,belle2018reasoning}, that are based on the \emph{Situation Calculus} ontology \cite{reiter2001knowledge}, can model imperfect sensors and effectors. The Situation Calculus' branching structure makes these frameworks mostly suitable for \textit{planning} under partial states of information. A recent extension of language $\mathcal{C}+$ \cite{giunchiglia2004nonmonotonic}, dubbed $p\mathcal{BC}+$, is presented in \cite{lee2018probabilistic}. The authors show how $p\mathcal{BC}+$ can be implemented in $\text{LP}^{\text{MLN}}$, a probabilistic extension of ASP that can be readily implemented using tools such as $\textit{LPMLN2ASP}$ \cite{lee2017computing}. Unlike our work, which focuses on runtime reasoning, $p\mathcal{BC}+$ is mostly suited for probabilistic diagnosis and parameter learning. The two languages \textit{MLN-EC} \cite{skarlatidis2015mlnec} and \textit{ProbEC} \cite{skarlatidis2015probabilistic} extend the semantics of the \emph{Event Calculus} \cite{kowalski1986logic,miller2002some} ontology, using Markov Logic Networks \cite{richardson2006markov} and ProbLog \cite{deraedt2007problog} to perform event recognition from security cameras. In their proposed case study, the logical part of the architecture receives time-stamped events as inputs and processes them in order to detect complex long-term activities (e.g., infer that two people are fighting from the fact that they have been close to each other and moving abruptly during the last few seconds). Given their semi-probabilistic nature, these frameworks are able to handle uncertainty in the input events (ProbEC) or in the causal rules linking events and fluents (MLN-EC) but they do not deal with any form of epistemic knowledge. On the other hand, EPEC \cite{dasaro2019phd} has the advantage of being based on the Event Calculus (making it particularly well suited for Event Recognition tasks) and being able to deal with epistemic aspects. Its semantics abstracts from any specific programming language and therefore it lends itself to task-specific optimizations, such as the runtime adaptation presented in this work.

\section{Conclusion and Future Work}\label{sec:Conlcusion}
    
    To summarize, the contribution of this paper is two-fold:
    \begin{itemize}
        \item It presents an architecture, currently being actively developed and tested, for the rehabilitation of children with Development Coordination Disorders. The suggested approach mixes machine learning, logic-based and gamification techniques.
        
        \item It introduces an novel implementation of a state-of-the-art probabilistic logic programming framework (EPEC) that can work at runtime.
    \end{itemize}
    
    The logical framework plays the role of an interface between the exergame and the machine-learning techniques, and is mainly used for two reasons: (i) guide the exergame according to a predefined strategy that is hard-encoded in the domain description, and (ii) provide human-understandable feedback after each therapeutic session, that can be used to log user performance over time.
    
    Nonetheless, there is much room for extending the scope and functionalities of our application. As we collect data about the users of our system, we can compare the decisions taken by the system to those that therapists and psychologists would take, and adapt our domain descriptions accordingly. In the future, we would like to do this semi-automatically by \textit{learning} the probabilities in our domain descriptions from data-streams annotated by experts.
    
    It is also possible to expand the feedback functionalities of our system. At the moment, feedback includes a series of graphs about the level of engagement of the patient and limited textual reports about his/her performance level (see Section \ref{sec:anExampleFromAvatea}). However, a full history of what happened throughout each therapeutic session is recorded in an EPEC-readable narrative. Implementing a device to automatically translate these narratives into natural language would greatly enhance the transparency of our system, and constitutes a future step of this work. In fact, ``explanations'' coming from our system are limited to plots such as the one depicted in Figure \ref{fig:avateaExample}, that shows what the engagement level of the child is throughout the therapeutic session, and reasons behind decisions are explained in terms of trespassed thresholds. Transforming them into natural language would greatly enhance intelligibility of our system.
    
\section*{Competing interests}
    The authors declare they were partially supported by MIUR within the POR Campania FESR 2014-2020 AVATEA ``Advanced Virtual Adaptive Technologies e-hEAlth'' research project.

\bibliographystyle{acmtrans}
\bibliography{bibliography}

\section*{Appendix A}\label{appendix:causalpartofdomain}
    This is the causal part of the domain (the i-proposition and c-propositions) for the example in Section \ref{sec:anExampleFromAvatea}. It assumes that the subject is initially fully engaged and doing the task correctly. An event of type $\LowAttention$ causes the level of $\Engagement$ to drop, whereas $\TaskCorrect$ is negatively affected by $\BadPosture$ and $\EyesNotFollowingTarget$ events. Since PEC-RUNTIME can also model concurrent actions, the effect of multiple events happening at the same time can also be modeled. In our scenarios, we assume that there are some (negative) synergies when multiple (distinct) events happen at the same time. Furthermore, when none of the previous activities is detected by the sensors, the value of $\Engagement$ and $\TaskCorrect$ grows.\\
    
    \noindent
    $\initiallyOneOf \{ (\{\lit{\Engagement}{\true},\lit{\TaskCorrect}{\true}\},1) \}$\\
    
    \noindent
    $\{ \lit{\EyesNotFollowingTarget}{\false}, \lit{\LowAttention}{\false}, \lit{\BadPosture}{\false} \}$\\
    \hspace*{2em}$\causesOneOf$\\
    \hspace*{4em}$\{ (\{\lit{\TaskCorrect}{\true},\lit{\Engagement}{\true}\},4/10),$\\
    \hspace*{4em}$(\{\},6/10) \}$\\
    \\
    \noindent
    $\{ \lit{\EyesNotFollowingTarget}{\true}, \lit{\LowAttention}{\false}, \lit{\BadPosture}{\false}, \lit{\Engagement}{\true} \}$\\
    \hspace*{2em}$\causesOneOf$\\
    \hspace*{4em}$\{ (\{\lit{\TaskCorrect}{\false},\lit{\Engagement}{\false}\},3/100),$\\
    \hspace*{4em}$(\{\lit{\TaskCorrect}{\false}\},17/100),$\\
    \hspace*{4em}$(\{\},8/10) \}$\\
    \\
    \noindent
    $\{ \lit{\EyesNotFollowingTarget}{\true}, \lit{\LowAttention}{\false}, \lit{\BadPosture}{\false}, \lit{\Engagement}{\false} \}$\\
    \hspace*{2em}$\causesOneOf$\\
    \hspace*{4em}$\{ (\{\lit{\TaskCorrect}{\false}\},3/10),$\\
    \hspace*{4em}$(\{\},7/10) \}$\\
    \\
    \noindent
    $\{ \lit{\EyesNotFollowingTarget}{\false}, \lit{\LowAttention}{\true}, \lit{\BadPosture}{\false} \}$\\
    \hspace*{2em}$\causesOneOf$\\
    \hspace*{4em}$\{ (\{\lit{\Engagement}{\false}\},3/10),$\\
    \hspace*{4em}$(\{\},7/10) \}$\\
    \\
    \noindent
    $\{ \lit{\EyesNotFollowingTarget}{\true}, \lit{\LowAttention}{\true}, \lit{\BadPosture}{\false} \}$\\
    \hspace*{2em}$\causesOneOf$\\
    \hspace*{4em}$\{ (\{\lit{\TaskCorrect}{\false},\lit{\Engagement}{\false}\},4/10),$\\
    \hspace*{4em}$(\{\},6/10) \}$\\
    \\
    \noindent
    $\{ \lit{\EyesNotFollowingTarget}{\false}, \lit{\LowAttention}{\false}, \lit{\BadPosture}{\true} \}$\\
    \hspace*{2em}$\causesOneOf$\\
    \hspace*{4em}$\{ (\{\lit{\TaskCorrect}{\false},\lit{\Engagement}{\false}\},3/100),$\\
    \hspace*{4em}$(\{\lit{\TaskCorrect}{\false}\},17/100),$\\
    \hspace*{4em}$(\{\},8/10) \}$\\
    \\
    \noindent
    $\{ \lit{\EyesNotFollowingTarget}{\true}, \lit{\LowAttention}{\false}, \lit{\BadPosture}{\true} \}$\\
    \hspace*{2em}$\causesOneOf$\\
    \hspace*{4em}$\{ (\{\lit{\TaskCorrect}{\false},\lit{\Engagement}{\false}\},1/10),$\\
    \hspace*{4em}$(\{\lit{\TaskCorrect}{\false}\},3/10),$\\
    \hspace*{4em}$(\{\},6/10) \}$\\
    \\
    \noindent
    $\{ \lit{\EyesNotFollowingTarget}{\true}, \lit{\LowAttention}{\true}, \lit{\BadPosture}{\true} \}$\\
    \hspace*{2em}$\causesOneOf$\\
    \hspace*{4em}$\{ (\{\lit{\TaskCorrect}{\false},\lit{\Engagement}{\false}\},3/10),$\\
    \hspace*{4em}$(\{\lit{\TaskCorrect}{\false}\},2/10),$\\
    \hspace*{4em}$(\{\},5/10) \}$\\
    
    \section*{Appendix B}\label{appendix:proofOfProposition}
    
    This appendix proves the following proposition\footnote{The notation used in this proof follows \cite{dasaro2017lpnmr} and \cite{dasaro2019phd}.}:
    
    \begin{proposition}
        Let $\mathcal{D}$ be any domain description such that $\mathcal{D} = \mathcal{D}_{\leq 0}$.
        Then, $M_{\mathcal{D}_{>0}}(\varphi) = M_{\mathcal{D}} (\varphi)$ for any i-formula $\varphi$ having instants $>0$.
    \end{proposition}
    
    \begin{proof}
    Since $\mathcal{D} = \mathcal{D}_{\leq 0}$ the entire narrative of $\mathcal{D}$ (possibly) occurs at $0$. Let $\varphi$ be a formula that only has instants $>0$. According to PEC's semantics,
    \begin{equation}\label{eq:proof}
    M_{\mathcal{D}}(\varphi) = \sum_{W \mmodels \varphi} i(W) \epsilon_{\mathcal{D}}(W) t_{\mathcal{D}}(W(0), W(>0))
    \end{equation}
    where $i(W)$ is the initial probability of the state in $W$, $\epsilon_{\mathcal{D}}(W)$ is the probability of occurrence of the narrative in $W$, and $t_{\mathcal{D}}(W(0),W(>0))$ is the probability of transition from the state $W(0)$ to $W(>0)$.
    
    Since the formula $\varphi$ only has instants $>0$ we can group together worlds by their state at $1$. That is, we construct equivalence classes $[W]$ such that $W'\in[W]$ if $W'(1)=W(1)$. Since $\mathcal{D}$ has no narrative occurring at instants $>0$, well-behaved worlds $W$ w.r.t. $\mathcal{D}$ are such that $W(I+1)=W(I)$ for all instants $I > 0$ due to persistence. Since $\varphi$ only has instants $>0$, if a world $W' \in [W]$ is such that $W' \mmodels \varphi$ then also $W \mmodels \varphi$, since $W$ and $W'$ can only possibly differ at $0$. Then, Equation (\ref{eq:proof}) continues as follows:
    \[
        = \sum_{[W]\mmodels \varphi} M_{\mathcal{D}_{>0}}([W]) = M_{\mathcal{D}_{>0}}(\varphi)
    \]
    by definition.
    \end{proof}
\end{document}

%% file: macros.tex
\DeclareRobustCommand{\mmodels}{\mathrel{||}\joinrel\Relbar}

\newcommand{\at}{\ensuremath{ \mbox{ \bf{at} }}}
\newcommand{\atsign}{\ensuremath{\mbox{{\bf @}}}}

\newcommand{\Attention}{\ensuremath{\mbox{\textit{Attention}}}}

\newcommand{\BadPosture}{\ensuremath{\mbox{\textit{BadPosture}}}}

\newcommand{\believes}{\ensuremath{\mbox{ \bf{believes} }}}

\newcommand{\causes}{\ensuremath{\mbox{ {\bf causes} }}}
\newcommand{\causesOneOf}{\ensuremath{\mbox{ {\bf causes-one-of} }}}

\newcommand{\Cfalse}{\ensuremath{\mbox{\textit{Cfalse}}}}
\newcommand{\Ctrue}{\ensuremath{\mbox{\textit{Ctrue}}}}

\newcommand{\currentIProp}{\ensuremath{\mbox{\textit{currentIProp}}}}
\newcommand{\currentNarrative}{\ensuremath{\mbox{\textit{currentNarrative}}}}

\newcommand{\Engagement}{\ensuremath{\mbox{\textit{Engagement}}}}

\newcommand{\EyesNotFollowingTarget}{\ensuremath{\mbox{\textit{EyesNotFollowingTarget}}}}

\newcommand{\False}{\bot}
\newcommand{\false}{\ensuremath{\mbox{\textit{false}}}}

\newcommand{\Flu}{\ensuremath{\mbox{\textit{Flu}}}}

\newcommand{\generateEventsFrom}{\ensuremath{\mbox{\textit{generateEventsFrom}}}}

\newcommand{\holdsWithProb}{\ensuremath{\mbox{ {\bf holds-with-prob }}}}

\newcommand{\ifBelieves}{\ensuremath{\mbox{ {\bf if-believes} }}}

\newcommand{\ifor}[2]{\ensuremath{[#1]@#2}}

\newcommand{\initiallyOneOf}{\ensuremath{\mbox{{\bf initially-one-of} }}}

\newcommand{\iprop}{\ensuremath{\mbox{\textit{iprop}}}}

\newcommand{\lit}[2]{\mbox{\ensuremath{ #1 \! = \! #2}}}

\newcommand{\LowAttention}{\ensuremath{\mbox{\textit{LowValence}}}}
\newcommand{\LowerDifficultyLevel}{\ensuremath{\mbox{\textit{LowerDifficultyLevel}}}}

\newcommand{\narr}{\textit{narr}}

\newcommand{\occursAt}{\ensuremath{\mbox{ {\bf occurs-at} }}}

\newcommand{\outputNarrative}{\ensuremath{\mbox{\textit{outputNarrative}}}}

\newcommand{\performedAt}{\ensuremath{\mbox{ {\bf performed-at} }}}

\newcommand{\PlaySound}{\ensuremath{\mbox{\textit{PlaySound}}}}

\newcommand{\senses}{\ensuremath{\mbox{ {\bf senses} }}}

\newcommand{\takesValues}{\ensuremath{\mbox{ {\bf takes-values} }}}

\newcommand{\TaskCorrect}{\ensuremath{\mbox{\textit{TaskCorrect}}}}

\newcommand{\Test}{\ensuremath{\mbox{\textit{Test}}}}

\newcommand{\TrackObject}{\ensuremath{\mbox{\textit{TrackObject}}}}

\newcommand{\true}{\ensuremath{\mbox{\textit{true}}}}
\newcommand{\True}{\top}

\newcommand{\withAccuracies}{\ensuremath{\mbox{ {\bf with-accuracies} }}}

\newcommand{\withProb}{\ensuremath{\mbox{ {\bf with-prob} }}}
\newcommand{\withProbs}{\ensuremath{\mbox{ {\bf with-probs} }}}